\documentclass{article}

\usepackage{arxiv}

\usepackage[utf8]{inputenc} % allow utf-8 input
\usepackage[T1]{fontenc}    % use 8-bit T1 fonts
\usepackage{hyperref}       % hyperlinks
\usepackage{url}            % simple URL typesetting
\usepackage{booktabs}       % professional-quality tables
\usepackage{amsfonts}       % blackboard math symbols
\usepackage{nicefrac}       % compact symbols for 1/2, etc.
\usepackage{microtype}      % microtypography
\usepackage{lipsum}
\usepackage{graphicx}
\usepackage{amsmath,amssymb,amsfonts}
\usepackage{textcomp}
\usepackage{booktabs}
\usepackage{multirow}
\usepackage{tabularx}

\usepackage[group-separator={,}]{siunitx}
\usepackage{natbib}
\usepackage{subcaption}
\usepackage{algorithm}
\usepackage[noend]{algpseudocode}

\newcommand{\sepnum}[1]{\num[group-separator={,}]{#1}}

\newcommand{\mycomment}[1]{}

\title{Scaling Federated Learning Solutions with Kubernetes for Synthesizing Histopathology Images}

\author{Andrei-Alexandru Preda\textsuperscript{1},
Iulian-Marius T\u{a}iatu\textsuperscript{1}, Dumitru-Clementin Cercel\textsuperscript{1}\thanks{Corresponding author: dumitru.cercel@upb.ro.} \\
\textsuperscript{1}National University of Science and Technology POLITEHNICA Bucharest, Bucharest, Romania \\
}

\begin{document}
\maketitle
\begin{abstract}
In the field of deep learning, large architectures often obtain the best performance for many tasks, but also require massive datasets. In the histological domain, tissue images are expensive to obtain and constitute sensitive medical information, raising concerns about data scarcity and privacy. Vision Transformers are state-of-the-art computer vision models that have proven helpful in many tasks, including image classification. In this work, we combine vision Transformers with generative adversarial networks to generate histopathological images related to colorectal cancer and test their quality by augmenting a training dataset, leading to improved classification accuracy. Then, we replicate this performance using the federated learning technique and a realistic Kubernetes setup with multiple nodes, simulating a scenario where the training dataset is split among several hospitals unable to share their information directly due to privacy concerns.
\end{abstract}

% keywords can be removed

\section{Introduction}\label{sec:introduction}

Lack of sufficient training data has always been a problem in training machine learning models~\citep{10.1145/3502287}. This problem has become more pronounced over time, as state-of-the-art models have become larger and more data-hungry, such as the Transformer~\citep{vaswani2017attention} family. Augmenting datasets means generating new training samples from existing ones using various techniques. Generative Adversarial Networks  (GANs)~\citep{goodfellow2014generative} are suitable for generating synthetic samples, especially synthetic images. These networks involve training two neural networks that compete against each other and have produced good results in the past~\citep{odena2017conditional,JOSE202143}.

In recent years, the public has become more aware of the importance of data privacy, especially in the medical and telecommunications fields. Federated learning~\citep{mcmahan2017communication} is a technique that allows training models in a non-centralized manner by distributing a model to independent clients, which then train it locally using their private data. In this way, training data do not have to leave the environment where they are safely stored and are prone to being shared. This technique is especially relevant in the medical domain, which deals with sensitive data~\citep{feki2021federated,10.1145/3412357}.

In this work, we experiment with several GAN variants (i.e.,  conditional GAN~\citep{mirza2014conditional}, Wasserstein GAN with gradient penalty~\citep{10.5555/3295222.3295327}, and auxiliary classifier GAN~\citep{odena2017conditional}) to produce histopathological images and build a system capable of augmenting training datasets. Since the medical domain involves working with sensitive information, we then try to address realistic privacy concerns by incorporating federated learning into the process, simulating a situation where multiple hospitals would train this machine learning model without risking exposing their patients' data. Finally, we use Kubernetes (\url{https://kubernetes.io}, accessed on 10 March 2025) to test the project in a production-ready environment. The performance of the resulting system is evaluated by analyzing the images it produces. In addition, we assess their usefulness by demonstrating their effectiveness in a downstream task: training a separate classifier for medical images and checking to improve its classification capabilities when adding our synthetic data.
%Kubernetes~\citep{k8swhat}

Our main contributions are as follows:
\begin{enumerate}
    \item Comparing several GAN architectures for synthesizing histopathology images of colorectal cancer;
    \item To the best of our knowledge, this is the first time the federated setting has been used to train GANs on the 
    TCGA-CRC-DX dataset~\citep{kather_2020_3832231} we use;
    \item Experimenting to replicate centralized performance in a production-ready federated setting;
    \item Analyzing data produced by GANs to check for common errors and suggest improvements;
    \item Publishing our experimental code (\url{https://github.com/pandrei7/paper-synthesizing-histopathology}, accessed on 10 March 2025) for the wider public to ease reproducibility.
\end{enumerate}
%\url{https://github.com/pandrei7/paper-synthesizing-histopathology}

% \footnote{\url{https://github.com/pandrei7/paper-synthesizing-histopathology}}
%%%%%%%%%%%%%%%%%%%%%%%%%%%%%%%%%%%%%%%%%%
%\section{Materials and Methods}

\section{Related Work}\label{sec:related-work}

\subsection{Generative Adversarial Networks}

GANs~\citep{goodfellow2014generative} refer to a framework for training neural networks capable of producing fake content with a distribution similar to that of real content. The framework uses an adversarial process to train two separate networks: a \textit{generator} and a \textit{discriminator}. The objective of the generator is to produce realistic samples when given noise from a fixed distribution as input, while the discriminator tries to discern between real and fake samples. The two networks compete throughout the training process until the discriminator can no longer distinguish between real and fake input.

Since their introduction, multiple variants of GAN have appeared. One of these variants is the auxiliary classifier GAN (ACGAN)~\citep{odena2017conditional}. Unlike the original GAN, the generator of the ACGAN is conditioned on class labels, which later allows the generation of data of multiple specified types. Similarly, the discriminator is tasked with detecting whether a sample is real or fake and predicting the sample's class label.

It should be noted that while GANs have been used in the past~\citep{mino2018logan,esser2021taming,fang2022dp,nuastuasescu2022conditional} with good results, they tend to be challenging to train, manifesting problems such as overfitting and mode collapse, which refers to the generation of a limited number of types of samples that are exploited by a discriminator weakness~\citep{xu2023vit}.

\subsection{Vision Transformers}

The Transformer architecture~\citep{vaswani2017attention} was introduced in 2017, initially for the task of machine translation, but has since found many other uses, especially in the domain of natural language processing~\citep{paraschiv2019upb, vlad2019sentence, tanase2020upb, vlad2020upb, vlad2020upbevalita, paraschiv2020upb, echim2023adversarial}. Some of its characteristics include the encoder-decoder architecture and an attention mechanism. The vision Transformer (ViT)~\citep{dosovitskiy2020image} brings these ideas to the computer vision domain, splitting images into patches, linearly embedding them, adding position embeddings, and feeding them to a standard Transformer. For example, ViT has been used in medical applications as a component of the MedViT models~\citep{manzari2023medvit}.

\subsection{Federated Learning}

Federated learning~\citep{mcmahan2017communication} is a machine learning technique that allows training models to use datasets distributed on multiple \textit{clients} to avoid sharing private information with a centralized \textit{server}. At a high level, this technique works by having each client train a model on their private dataset and sending these models or only their gradients to the server, where they are aggregated. In this way, the training data never have to leave the clients. The server is in charge of coordinating the process until the training is over.

Federated learning is essential in the medical domain, where hospitals or research centers tend to have sensitive data that should not be shared, for example, for ethical reasons.

\subsection{Histopathology Image Analysis and Generation}

Histopathology is the study of tissue disease. An application of machine learning related to histopathology is the diagnosis of diseases by studying images of tissues and cells, which can be formulated as a classification problem. Histopathological images are more diverse, complex, and intricate than natural images, making this task difficult~\citep{fang2022dp}. In addition, the datasets for this task tend to be smaller and imbalanced, making generating synthetic images quite important. Our work is inspired by previous research that attempted to solve this problem. For example, \citep{xu2023vit} uses a diffusion model~\cite{ho2020denoising} enhanced with a Transformer to generate histopathological images. In contrast, \citep{zedda2023hierarchical} incorporates a hierarchical ViT~\citep{dosovitskiy2020image} into a custom architecture called a Hierarchical Pretrained Backbone Vision Transformer to classify histopathological images.

\section{Method}\label{sec:methods}

%These experiments are summarized in \Cref{fig:overview}.

\begin{figure*}[tb]
    \centering
    \includegraphics[width=\textwidth]{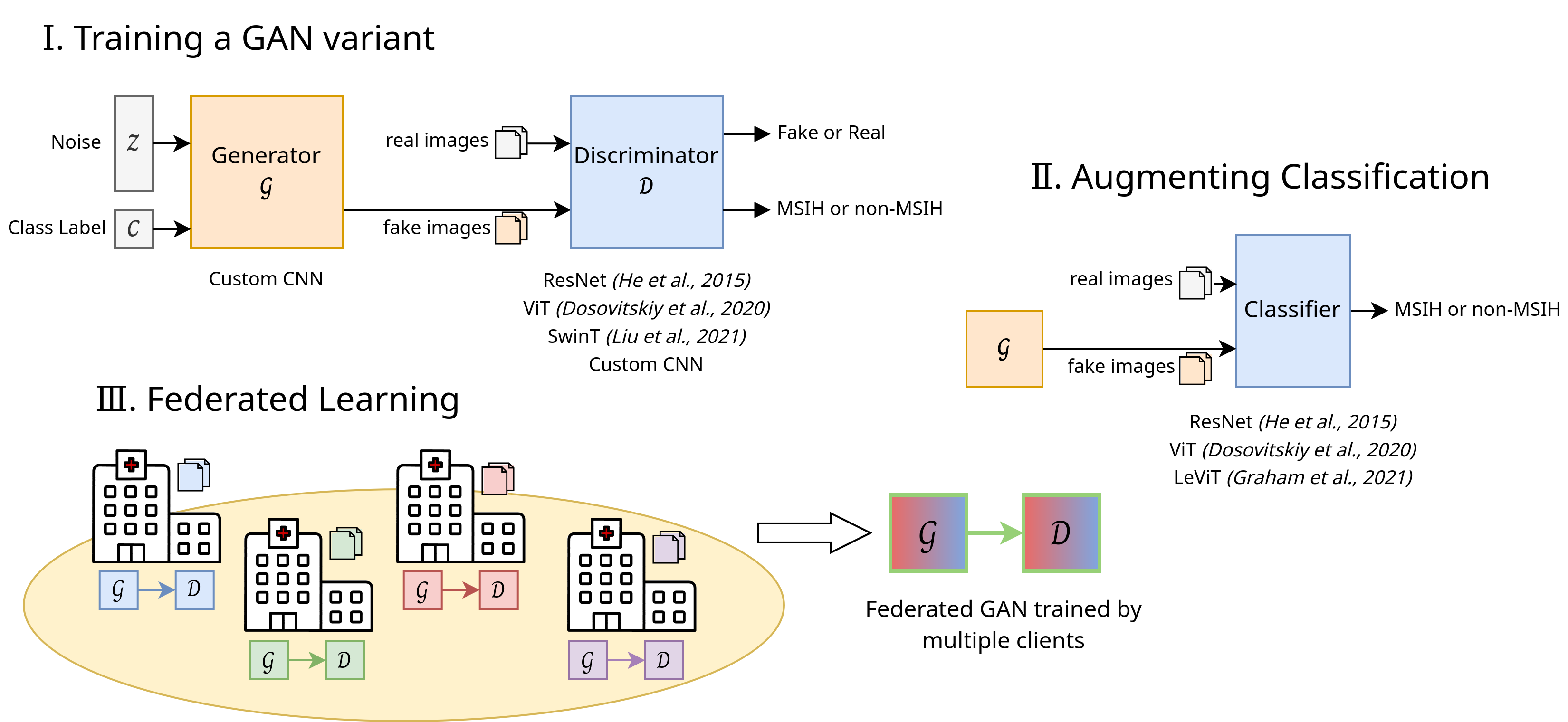}
    \caption{Summary of the proposed approach: testing different types of GANs for histopathology images, using synthetic images to improve classification, and exploring this method in a federated setting for increased data privacy.}
    \label{fig:overview}
\end{figure*}

\subsection{Data Augmentation}

\subsubsection{GAN-based Augmentations}\label{sec:gans}

As shown in Figure \ref{fig:overview}, our solutions for generating synthetic images are based on different
variants of the discriminator network. Specifically, we started by experimenting with a convolutional neural network (CNN) \cite{o2015introduction}, namely ResNet~\citep{he2016deep}, and then moved to Transformer architectures, such as the original ViT~\citep{dosovitskiy2020image} and the Swin Transformer (SwinT)~\citep{Liu2021SwinTH}, which obtained better results.

%We experimented with multiple architectures for the discriminator and obtained the best results with more robust networks based on Transformers.

In all experiments, the generator is a \textit{Custom CNN}, consisting of a convolutional neural network with three layers, the LeakyReLU activation function, and operations such as batch normalization, upsampling, and dropout. To control the labels of synthetic images, we condition the generator on the class embeddings learned throughout the training. These embeddings can be added as an additional dimension to the noise tensor. However, we obtained better results by multiplying each input channel with the embedding instead; our embeddings were shaped like image channels.

There are multiple ways of training GANs. We experimented with the regular conditional GAN (CGAN)~\citep{mirza2014conditional}, the Wasserstein GAN~\citep{pmlr-v70-arjovsky17a} variant that adds a gradient penalty to stabilize training (WGAN-GP)~\citep{10.5555/3295222.3295327}, and with the auxiliary classifier GAN (ACGAN)~\citep{odena2017conditional}, which requires the discriminator to guess both the image label and its origin.

\subsubsection{Geometric Transformations}

% We tested both architectures based on convolutional neural networks (CNNs) \cite{kim2014convolutional} and those based on Transformers (i.e., ViT~B~16, ViT~L~16~\citep{dosovitskiy2020image} and LeViT Conv 384~\citep{Graham2021LeViTAV}). (This is mentioned in "Classification Models" and does not fit here anymore.)

To train better classifiers, we use data augmentation through geometric transformations: crops, horizontal flips, and affine transformations up to at most 45 degrees. In addition, as a data preprocessing step, all pixel values are normalized. These geometric transformations are also applied to the synthetic images generated with GANs.

\subsection{Classification Models}
One way we tried to evaluate the quality of the synthetic images produced by GANs was to check their usefulness for classification through augmentation. We handle classification starting from pre-trained models and changing their classification heads before fine-tuning them on our dataset. We used multiple classifiers with different internal mechanisms:
\begin{itemize}
    \item ResNet~\citep{he2016deep}: ResNet is a popular neural network architecture for computer vision and is one of the first to demonstrate the effectiveness of residual connections in training deeper models. This architecture processes images using convolutions, residual blocks, and downsampling operations. We specifically used the \textit{ResNet 50} variant~\citep{torchvision2016} in our experiments, which contains five different stages, each containing several convolutional and batch normalization layers, as well as the ReLU activation function. This classifier was initialized with weights pre-trained on the ImageNet dataset~\citep{deng2009imagenet}.
    \item ViT~\citep{dosovitskiy2020image}: The ViT architecture incorporates the attention mechanism into a solution for computer vision by splitting images into several parts and applying attention to their embeddings. These models use several Transformer encoder layers, involving both attention operations and fully connected layers, and using the GELU activation function. We specifically used the \textit{ViT~B~16} and \textit{ViT~L~16} variants present in TorchVision~\citep{torchvision2016}, with 86 million trainable parameters and 305 million, respectively. These classifiers were also pre-trained on ImageNet before our fine-tuning.
        \item LeViT~\citep{Graham2021LeViTAV}: The LeViT architecture is a hybrid neural network that combines convolutions and attention. The input is first processed by a series of convolutional layers, then by three stages of modified Transformer blocks. We specifically used the implementation available in the timm library~\citep{rw2019timm}, where the model is called \textit{LeViT Conv 384}. This model had approximately 37 million trainable parameters, and, different from the previous two classifiers, it uses the Hardswish activation function.
\end{itemize}

\subsection{Federated Learning}

After obtaining the results in a centralized setting, we try to replicate that performance using federated learning to simulate multiple hospitals that collaborate to train a good model while keeping their data private. Concretely, we perform numerous rounds of training, which involve sampling a few clients, sending them the current aggregate model checkpoint, improving that model for a few epochs separately on each client, and then merging the updated weights to obtain the next aggregate model. We perform weighted averaging using the FedAvg algorithm~\citep{mcmahan2017communication}. The Algorithm \ref{alg:fl-pseudo} shows the pseudocode.

\begin{algorithm}
	\caption{
		Pseudocode for the federated learning algorithm using FedAvg~\citep{mcmahan2017communication}\\
		$R =$ number of rounds to train\\
		$E =$ number of epochs to train a client during one round\\
		$\alpha =$ fraction of clients to train in one round\\
		$\delta_i =$ dataset belonging to client $i$ \\
		$w_{c,i} =$ weight coefficient computed by FedAvg
	}
	\begin{algorithmic}[1]
		\Procedure{Training}{}
		\State $\mathrm{model_0} \gets \textit{random GAN initialization}$
		\For{$i = 1 \to R$}
		\State {$\mathrm{clients} \gets$ sample $\alpha$ client IDs}
		\ForAll {$c \in \mathrm{clients}$}
		\State {$\mathrm{weights_c} \gets \Call{Round}{\mathrm{model_{i-1}},\delta_c}$}
		\EndFor
		\State $\mathrm{model_i} \gets \sum_{c \in \mathrm{clients}}{w_{c,i} * \mathrm{weights_c}}$
		\EndFor
		\State \Return $\mathrm{model_R}$
		\EndProcedure
		% \State $ $
        \vspace{.5\baselineskip}

		\Procedure{Round}{$\mathrm{model_0}$, dataset}
		\For{$i = 1 \to E$}
		\State $\mathrm{model_i} \gets \Call{Backpropagation}{\mathrm{model_{i-1}},\mathrm{dataset}}$
		\EndFor
		\State \Return $\mathrm{model_E}$
		\EndProcedure
	\end{algorithmic}
 \label{alg:fl-pseudo}
\end{algorithm}

In practice, different clients often have different data distributions, which can affect the training process \cite{feki2021federated}. To simulate this scenario, we introduce a class imbalance in every client by assigning more samples from one of the two classes. We call this scenario \textit{the not independent and identically distributed (non-IID)} \cite{feki2021federated} and study the effect of the class imbalance rate.

\subsection{Experimental Setup}\label{sec:experiments}

\subsubsection{Dataset}
%``MSIH''
We experimented with the TCGA-CRC-DX~dataset~\citep{kather_2020_3832231}, which consists of histopathological images \sepnum{51918} of colorectal cancer. These belong to patients divided into two categories based on the microsatellite instability (MSI) status: samples from patients with MSI-high form class \textit{MSIH}, and those from patients with MSI-low or with microsatellite stability (MSS) form the \textit{non-MSIH} class. Thus, the dataset is suitable for a binary classification task.

The original images had a resolution of $512\times512$, but we resized them to $224\times224$ to lower the computing resources required. We used the train-test split published on Zenodo (\url{https://zenodo.org/records/3832231}, accessed on 10 March 2025), which includes \sepnum{19557} images in the training set and \sepnum{32361} images in the test set, a ratio of $1:2$.

\subsubsection{Hyperparameters}

Each variant of GAN was trained for $20$--$40$ epochs, with a learning rate of $10^{-4}$ (or $3*10^{-5}$ for ACGAN with a SwinT discriminator). In general, training the GANs well was fairly difficult and required setting the right hyperparameters. WGAN-GP especially proved to be sensitive to hyperparameters such as $n_{critic}$ (which controls the number of epochs used to train the discriminator) and $\lambda_{GP}$ (the coefficient for gradient penalty). Setting improper values for these parameters resulted in completely noisy images and an ineffective training process. We used $n_{critic}=10$ and $\lambda_{GP}=3$. To search for GAN hyperparameters, we typically started several training runs with various reasonable combinations and stopped them after a small number of epochs. Then, based on the stability of the loss curves and manual inspection of the images produced so far, we trained much longer with the most promising configurations. Although this simple method can be improved, in practice, it helped us find good hyperparameters quickly.

For classification, models were trained for $30$--$50$ epochs with early stopping and a learning rate of $10^{-4}$. The images were normalized with the mean and standard deviation suitable for ResNet. The best models were initialized from weights pre-trained on ImageNet.

For federated learning experiments, we trained with a number of $4$ or $10$ clients for between $10$--$16$ rounds, each round involving the choice of $4$ random clients and training them for $4$ local epochs; thus, the fraction of clients $\alpha$ mentioned in Algorithm \ref{alg:fl-pseudo} was $40\%$ or $100\%$, depending on the run.

Experiments were carried out using Python libraries related to PyTorch~\citep{paszke2019pytorch} as follows: Torchvision~\citep{torchvision2016}, PyTorch Lightning~\citep{Falcon_PyTorch_Lightning_2019}, and Torchmetrics~\cite{Nicki_Skafte_Detlefsen_and_Jiri_Borovec_and_Justus_Schock_and_Ananya_Harsh_and_Teddy_Koker_and_Luca_Di_Liello_and_Daniel_Stancl_and_Changsheng_Quan_and_Maxim_Grechkin_and_William_Falcon_TorchMetrics_-_Measuring_2022}. Federated learning experiments used the Flower framework~\citep{beutel2022flower} for Python.
%, and its implementation of the Federated Averaging (FedAvg)~\citep{mcmahan2017communication} algorithm.

\subsubsection{Baselines}

To prepare for future experiments, we first set a baseline classification performance for some models on the original dataset. Specifically, we tested four different models: ResNet~50, ViT~B~16, ViT~L~16, and LeViT~Conv~384. The results can be seen in Table \ref{tab:classification}. ViT~L~16 obtained the best F1-score on this dataset, close to $88\%$.

\begin{table}[tb]
	\caption{Baseline performance on the test set. Augmentation refers to geometric transformations such as crops, flips, and affine transformations.}
	\label{tab:classification}
	\centering
	\begin{tabular}{l c c c}
		\toprule
		Model          & Augmentation & Accuracy         & F1-score         \\
		\midrule
		ResNet 50~\citep{he2016deep,torchvision2016}      & No                  & $.7965         $ & $.8800         $ \\
		ResNet 50~\citep{he2016deep,torchvision2016}      & Yes                 & $.7778         $ & $.8663         $ \\
		ViT B 16~\citep{dosovitskiy2020image,torchvision2016}       & Yes                 & $.7952         $ & $.8802         $ \\
		ViT L 16~\citep{dosovitskiy2020image,torchvision2016}       & Yes                 & $\mathbf{.8005}$ & $\mathbf{.8833}$ \\
		LeViT Conv 384~\citep{Graham2021LeViTAV,rw2019timm} & Yes                 & $.7846         $ & $.8713         $ \\
		\bottomrule
	\end{tabular}
\end{table}

%%%%%%%%%%%%%%%%%%%%%%%%%%%%%%%%%%%%%%%%%%
\section{Results}

\subsection{GAN Comparison}

FID scores~\citep{DBLP:journals/corr/HeuselRUNKH17} obtained with different GANs and discriminators can be seen in Table \ref{tab:fid-discriminators}.  We observed stark differences when training with a CNN-based discriminator instead of a Transformer-based discriminator, with Transformer discriminators giving better results. More powerful discriminators were able to learn relevant features about the inputs, which in turn helped the generator. The attention mechanism employed by ViT, combined with the pre-trained initialization weights that identify visual features fit for image classification, gave it an edge in processing histopathological images, which are very intricate. Similarly, having different architectures for the generator and discriminator meant that we encountered fewer situations where one component of the GAN exploited a weakness of the other, leading the system to diverge, as happened more often when using only CNNs.  

\begin{figure*}[tb]
	\centering
	\begin{subfigure}[b]{.32\textwidth}
		\includegraphics[width=\textwidth]{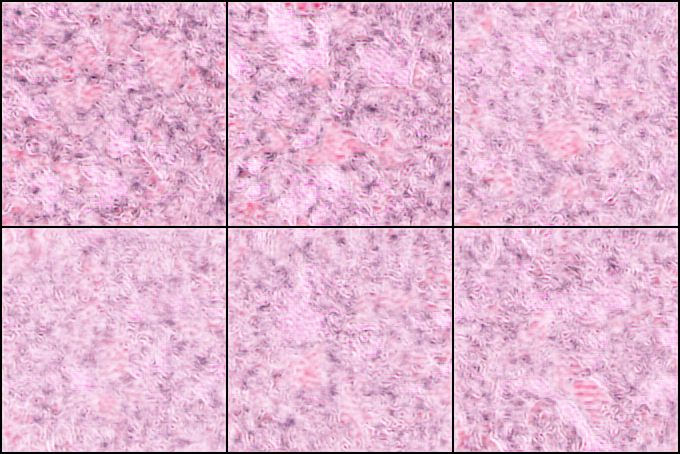}
		\caption{\centering After 5 epochs\vspace{8pt}}
	\end{subfigure}
	\begin{subfigure}[b]{.32\textwidth}
		\includegraphics[width=\textwidth]{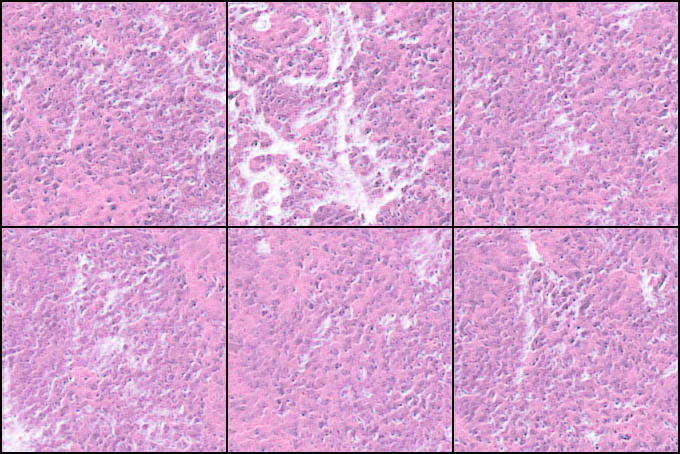}
		\caption{\centering After 12 epochs\vspace{8pt}}
	\end{subfigure}
	\begin{subfigure}[b]{.32\textwidth}
		\includegraphics[width=\textwidth]{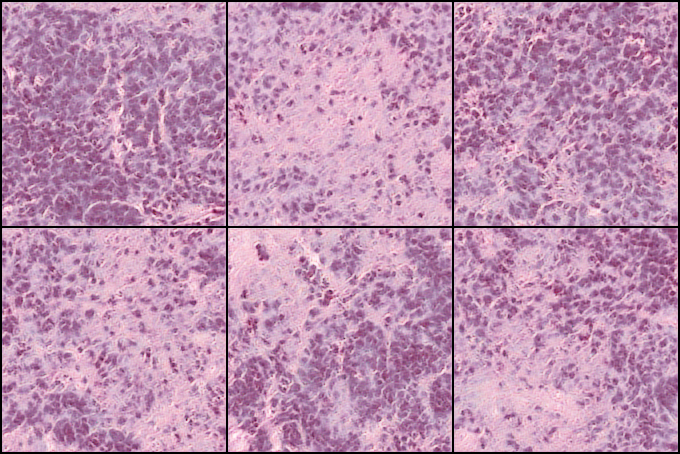}
		\caption{\centering After 19 epochs\vspace{8pt}}
	\end{subfigure}
	\begin{subfigure}[b]{.32\textwidth}
		\includegraphics[width=\textwidth]{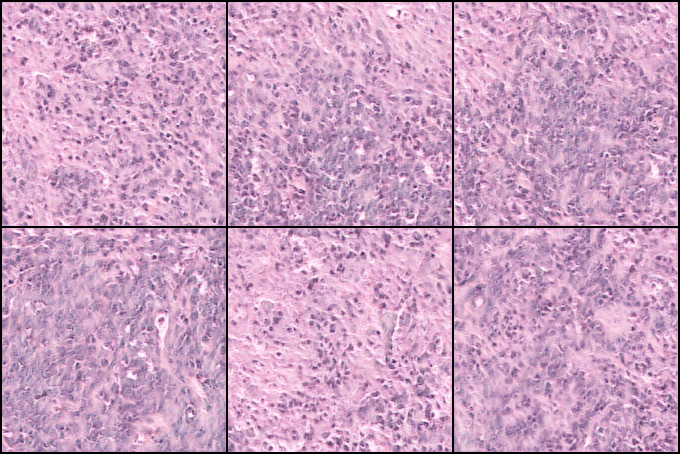}
		\caption{\centering After 26 epochs\vspace{8pt}}
	\end{subfigure}
	\begin{subfigure}[b]{.32\textwidth}
		\includegraphics[width=\textwidth]{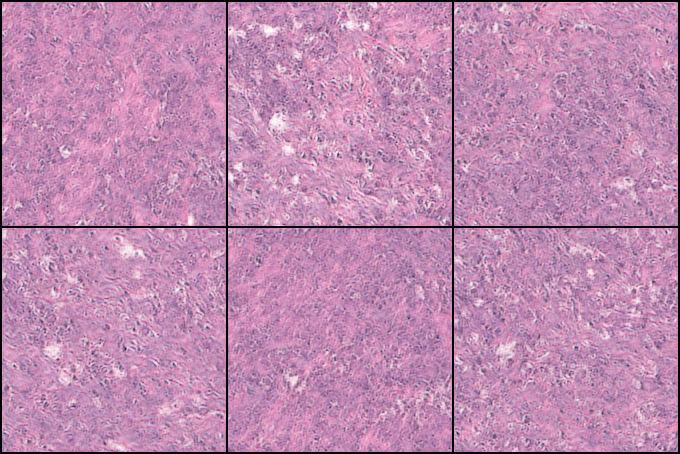}
		\caption{\centering After 33 epochs\vspace{8pt}}
	\end{subfigure}
	\begin{subfigure}[b]{.32\textwidth}
		\includegraphics[width=\textwidth]{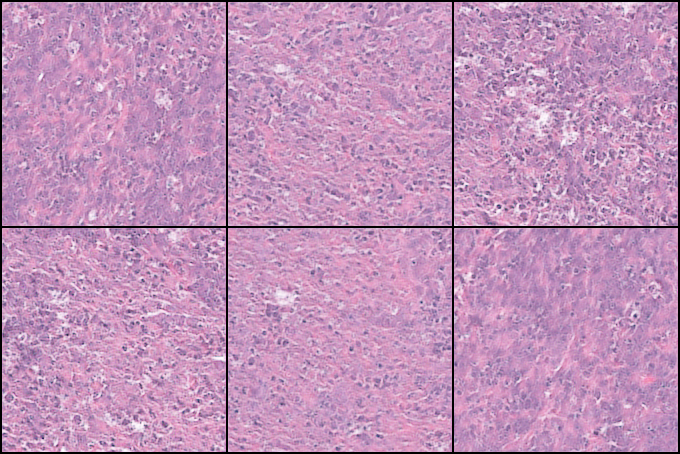}
		\caption{\centering After 40 epochs\vspace{8pt}}
	\end{subfigure}
	\caption{Images generated by ACGAN with a ViT B 16 discriminator.}
	\label{fig:evolution}
\end{figure*}

Furthermore, architectures such as the CGAN had a harder time producing images with any amount of detail, with their output usually being noisy color gradients. The extra classification task imposed by ACGAN allowed the discriminator to develop better features, allowing the generator to produce images with more detail. Figure \ref{fig:evolution} displays the images produced by ACGAN during a training run using a ViT B 16 discriminator. 
As seen through visual inspection in Figure \ref{fig:evolution}, these results also reinforce the idea that image quality fluctuates during training. However, it should be noted that images with the lowest FID did not always look reasonable on visual inspection and cannot be judged based solely on this metric.

\begin{table*}[tb]
    \captionsetup{width=\textwidth}
	\caption{FID scores for different GANs and discriminators (lower is better)\protect\footnotemark.}
	\label{tab:fid-discriminators}
	\centering
	\begin{tabular*}{.8\textwidth}{@{\extracolsep\fill}l l c r}
		\toprule
		GAN Type & Discriminator & No. of Epochs & FID ($\downarrow$) \\
		\midrule
        ACGAN   & Custom CNN  & 30 & 355.30 \\
        % \midrule
                & ResNet 18   & 40 & 261.86             \\
		          & ResNet 50   & 40 & \textbf{143.91}    \\
		% \midrule
		        & ViT B 16    & 28 & 201.22             \\
		          &             & 32 & 223.86             \\
		          &             & 36 & 177.00    \\
		          &             & 40 & 192.29             \\
		% \midrule
                & SwinT        & 40 & 285.71             \\
	    \midrule
    	CGAN    & Custom CNN  & 30 & 333.97 \\
                & ViT B 16    & 30 & 325.36 \\
        \midrule
        WGAN-GP & Custom CNN  & 30 & 281.47 \\
        \bottomrule
	\end{tabular*}
\end{table*}
\footnotetext{For the ViT B 16 model, we evaluate image quality at multiple points during training to get a sense of how it changes in time (all these scores come from a single, continuous training run).}

\subsection{Augmented Classification}

In Table \ref{tab:augmented-classification}, we can find classification scores after augmenting the training dataset with images generated by ACGANs. In Table \ref{tab:augmentedothergans} are the results of the other configurations. Each of these runs included \sepnum{10000} synthetic images, and half of them also used the real training set, bringing the total number of training samples to \sepnum{29557} images. We observe that classifiers with architectures more like the discriminator benefit from synthetic images. This could happen because the generated images encode the kind of information that architecture exploits best. This might explain, at least partially, why sometimes using only synthetic data helped more than the combination of real and synthetic images. 
%(alongside the fact that there seems to be a somewhat high variance between different training runs). 

For ACGAN, SwinT is the discriminator that produced the most successful generator to improve classification performance, obtaining an F1-score of $90.43\%$. The overall best F1-score was obtained with data generated from the CGAN + ViT B 16 combination. Both of these high scores were obtained in the synthetic-only case, but the corresponding mixes with real data also improved the baseline.

\begin{table*}[tb]
    % \begin{adjustwidth}{-\extralength}{0cm}
    \centering
	\caption{Results for different classifiers with synthetic data from ACGAN. All synthetic images were generated by a CNN trained using an ACGAN, but with different discriminators. Improvements over the baseline (no synthetic data) are highlighted in bold.}
	\label{tab:augmented-classification}
	% \begin{tabular*}{.8\textwidth}{{@{\extracolsep\fill}cccccccccc}}
        \begin{tabularx}{\textwidth}{c c c *{8}{>{\centering\arraybackslash}X}}
		\toprule
		ACGAN                                          & Data     & \multicolumn{2}{c}{ResNet 50} & \multicolumn{2}{c}{ViT B 16} & \multicolumn{2}{c}{ViT L 16} & \multicolumn{2}{c}{LeViT Conv 384}                                                   \\
		Discriminator                                  & Type        & Acc.                           & F1                            & Acc.                          & F1                                 & Acc.  & F1    & Acc.           & F1             \\ \midrule
		- & only real & .7965       & .8800                          & .7952                         & .8802                         & .8005                              & .8833 & .7846 & .8713                           \\ \midrule
		ResNet 18                           & only generated   & \textbf{.8052}                 & \textbf{.8910}                & .6717                         & .7979                           & .6684 & .7890 & \textbf{.8102} & \textbf{.8949} \\
		    @40 epochs                                           & generated + real & .7866                          & .8718                         & .7870                         & .8720                              & .7985 & .8816 & \textbf{.8078} & \textbf{.8891} \\ %\cline{3-10}
		ResNet 50                           & only generated   & .7462                          & .8437                         & .7410                         & .8431                              & .7456 & .8451 & .5727          & .6897          \\
		        @40 epochs                                       & generated + real & .7777                          & .8660                         & .7907                         & .8731                              & .7901 & .8736 & .7806          & .8698          \\ \midrule
		ViT B 16                                 & only generated   & \textbf{.8167}                 & \textbf{.8979}                & .4694                         & .5853                              & .5030 & .6243 & .7679          & .8664          \\
		       @28 epochs                                        & generated + real & .7763                          & .8651                         & .7907                         & .8745                              & .7926 & .8774 & .7641          & .8580          \\ %\cline{3-10}
		ViT B 16                               & only generated   & .4318                          & .5382                         & .3356                         & .3954                              & .3884 & .4789 & .3545          & .4056          \\
		       @32 epochs                                        & generated + real & .7903                          & .8755                         & .7847                         & .8711                              & .7893 & .8765 & .7814          & .8668          \\ %\cline{3-10}
		ViT B 16                                 & only generated   & .4432                          & .5521                         & .5002                         & .6237                              & .5152 & .6417 & .3901          & .4724          \\
		        @36 epochs                                       & generated + real & .7842                          & .8718                         & .7908                         & .8768                              & .7755 & .8638 & \textbf{.7916} & \textbf{.8763} \\ %\cline{3-10}
		ViT B 16                                 & only generated   & .4333                          & .5301                         & .7605                         & .8594                              & .5215 & .6479 & .6296          & .7549          \\
		        @40 epochs                                       & generated + real & .7730                          & .8619                         & .7798                         & .8667                              & .7852 & .8731 & \textbf{.7957} & \textbf{.8800} \\ \midrule
		SwinT                                & only generated   & \textbf{.8055}                 & \textbf{.8910}                & \textbf{.8257}                & \textbf{.9043}                     & .7748 & .8722 & \textbf{.8225} & \textbf{.9025} \\
		      @40 epochs                                         & generated + real & \textbf{.8063}                 & \textbf{.8869}                & .7842                         & .8704                              & .7958 & .8808 & .7840          & \textbf{.8716} \\
        \bottomrule
	\end{tabularx}
    % \footnotesize{All synthetic images were generated by a CNN trained using an ACGAN, but with different discriminators. Highlighted in bold are improvements over the baseline (no synthetic data).}
    % \end{adjustwidth}
\end{table*}

\begin{table*}[tb]
	% \begin{adjustwidth}{-\extralength}{0cm}
    \centering
	\caption{Results for different classifiers with synthetic data from various GANs. All synthetic images were generated by a CNN trained using GANs other than ACGAN. Improvements over the baseline (no synthetic data) are highlighted in bold.}
	\label{tab:augmentedothergans}
    % \begin{tabular*}{\textwidth}{{@{\extracolsep\fill}ccccccccccc}}
    \begin{tabularx}{\textwidth}{c c c *{8}{>{\centering\arraybackslash}X}}
    \toprule
GAN              & GAN Discriminator & \multicolumn{1}{c}{Data}     & \multicolumn{2}{c}{ResNet 50}  & \multicolumn{2}{c}{ViT B 16}   & \multicolumn{2}{c}{ViT L 16} & \multicolumn{2}{c}{LeViT Conv 384} \\
Type      & & \multicolumn{1}{c}{Type}                             & Acc.           & F1             & Acc.           & F1             & Acc.          & F1            & Acc.             & F1              \\
\midrule 
- & - & only real        & .7965          & .8800          & .7952          & .8802          & .8005         & .8833         & .7846            & .8713           \\
\midrule
CGAN & Custom CNN        & \multicolumn{1}{c}{only generated}   & .7379          & .8465          & .7483          & .8520          & .5463         & .6787         & .5720            & .7117           \\
     &              & \multicolumn{1}{c}{generated + real} & .7939          & .8780          & .7695          & .8616          & .7789         & .8689         & \textbf{.7874}   & \textbf{.8737}  \\
CGAN & ViT B 16 & \multicolumn{1}{c}{only generated}   & \textbf{.8363} & \textbf{.9108} & \textbf{.8216} & \textbf{.9013} & .6362         & .7576         & \textbf{.8395}   & \textbf{.9124}  \\
     &         & \multicolumn{1}{c}{generated + real} & .7938          & .8784          & .7898          & .8753          & .7717         & .8620         & \textbf{.7916}   & \textbf{.8752}  \\
WGAN-GP & Custom CNN & \multicolumn{1}{c}{only generated}   & .4445          & .5497          & .7188          & .8317          & .6615         & .7896         & .6142            & .7422           \\
     &        & \multicolumn{1}{c}{generated + real} & .7957          & .8784          & .7895          & .8759          & .7951         & .8783         & .7845            & .8718          \\
    \bottomrule
\end{tabularx}
% \end{adjustwidth}
% \noindent{\footnotesize{All synthetic images were generated by a CNN trained using GANs other than ACGAN, in different configurations. Highlighted in bold are improvements over the baseline (no synthetic data).}}
\end{table*}

\subsection{Federated Learning}

%In the medical setting, privacy is very important, and this is reflected in patient data confidentiality. Different medical facilities are usually not allowed to share their patient data, which means that federated learning techniques are often required. Even more, different medical facilities tend to have different data distributions, which is a problem that we have to tackle to train good models.

\begin{figure*}[tb]
 \centering
 \begin{subfigure}[b]{.32\textwidth}
  \includegraphics[width=\textwidth]{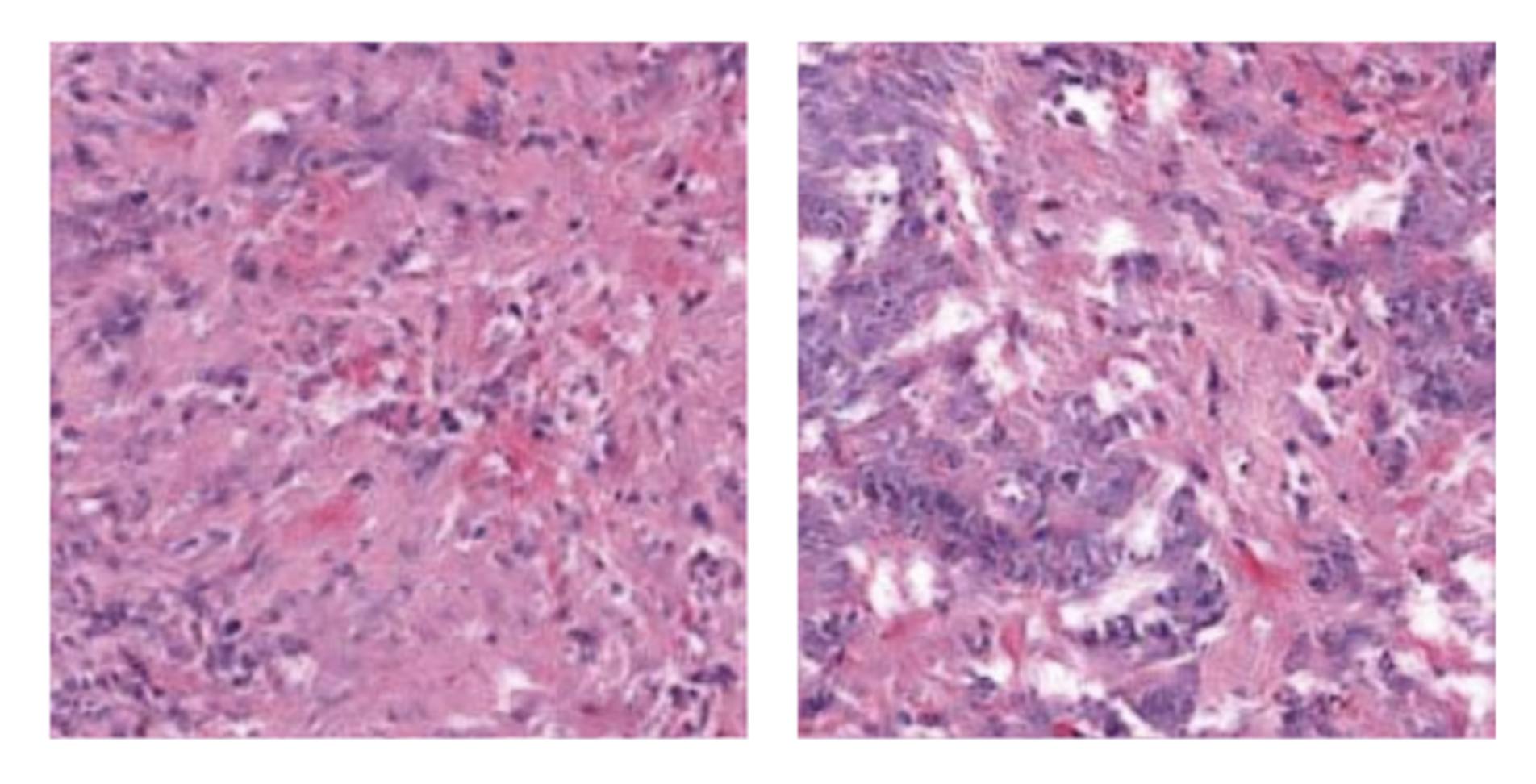}
  \vspace{-10pt}
  \caption{\centering Centralized\vspace{8pt}}
 \end{subfigure}
  \begin{subfigure}[b]{.32\textwidth}
  \includegraphics[width=\textwidth]{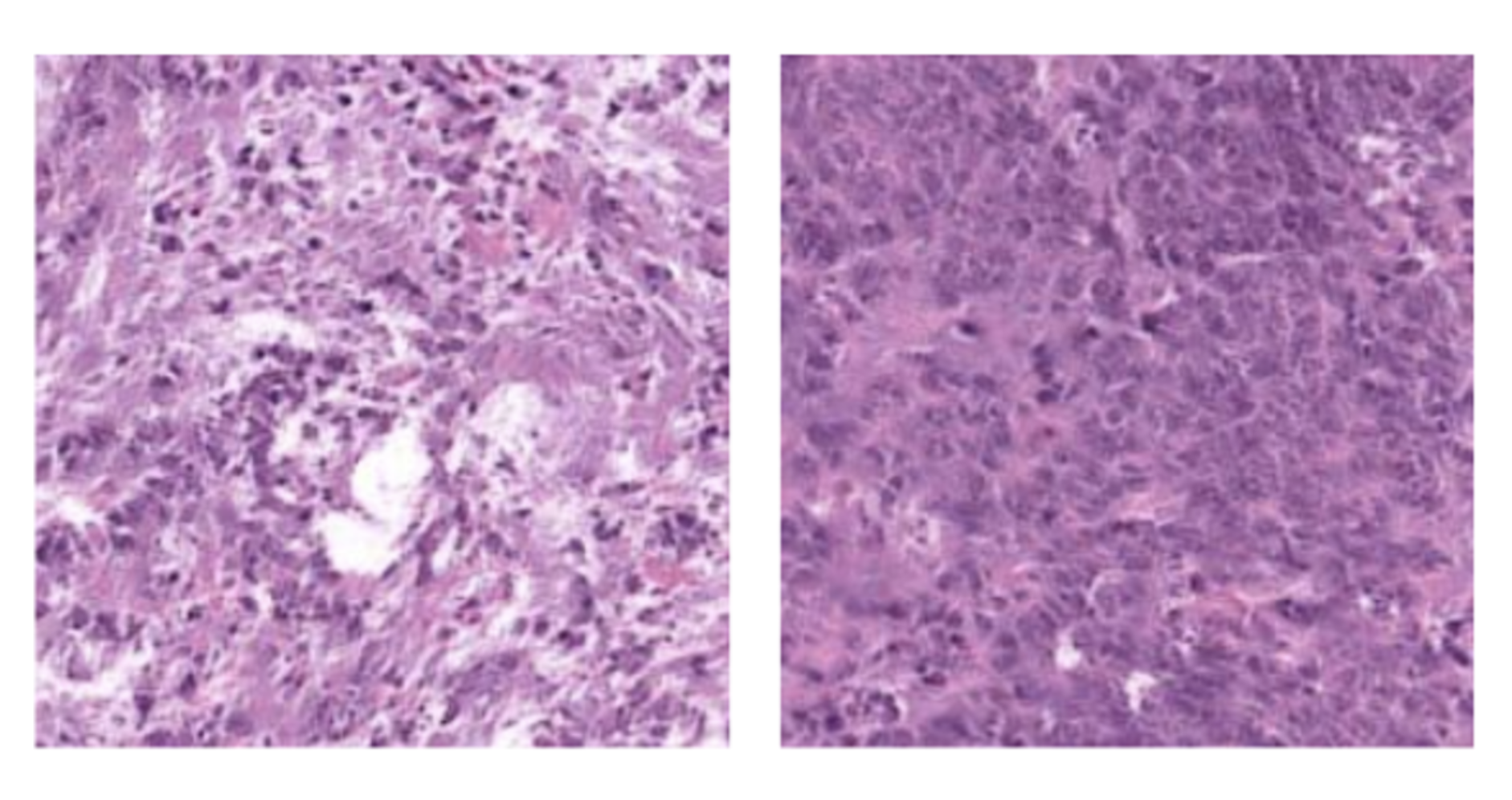}
  \vspace{-11pt}
  \caption{\centering 4 clients\vspace{8pt}}
 \end{subfigure}
 \begin{subfigure}[b]{.32\textwidth}
  \includegraphics[width=\textwidth]{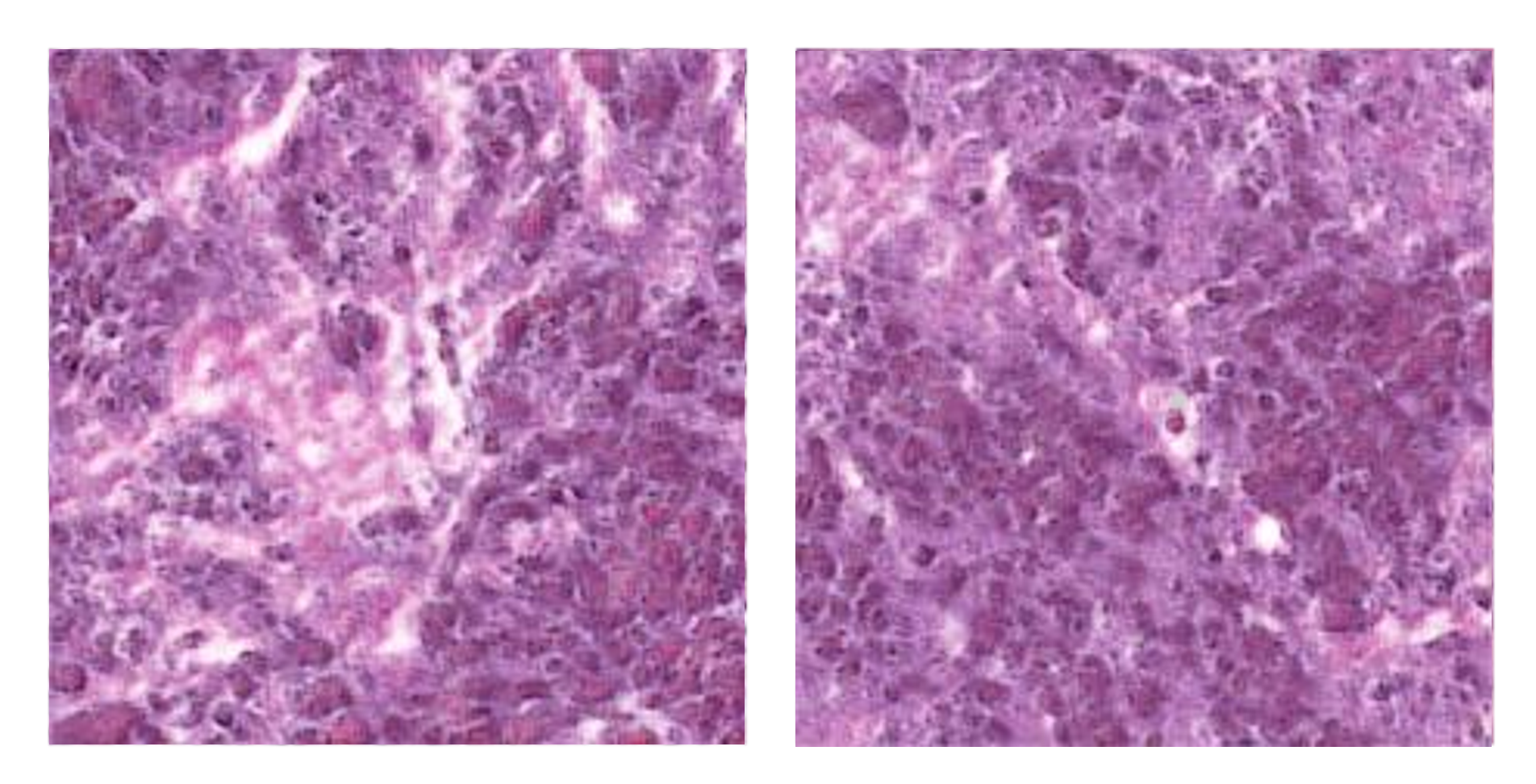}
  \vspace{-10pt}
  \caption{\centering 10 clients\vspace{8pt}}
 \end{subfigure}
 \caption{Comparison of images generated in the centralized vs. federated settings.}
 \label{fig:centralized-vs-federated}
\end{figure*}

% We also analyzed the effect of non-independent and identically distributed (non-IID) client data on performance.
%Inspired by \citet{feki2021federated},
Here, we tried to replicate the performance of the centralized GAN experiments, this time in a federated environment. We can find classification scores with federated learning-augmented training data in Table \ref{tab:classif-fl}. Because the number of clients affects the training time, the stability of the training, and overall performance, we analyzed two settings: one in which the data was split among four different clients and another with even more fragmentation, using ten clients. In both cases, the final classification scores were similar to the centralized setting, suggesting that this task adapted well to the federated setting. Similarly, FID scores (see Table \ref{tab:fid-fl}) with federated learning ranged from $163.86$ to $245.56$, which is a similar range of values to their centralized variants. Finally, we can find a visual comparison of the two types of images in Figure \ref{fig:centralized-vs-federated}.

%\multicolumn{2}{c}{Original w/o generated}
\begin{table*}[tb]
	% \begin{adjustwidth}{-\extralength}{0cm}
    \centering
	\caption{Classification scores when augmenting the training dataset with synthetic data from generators trained in the federated setting. Improvements over the baseline (no synthetic data) are highlighted in bold.}
	\label{tab:classif-fl}
	% \begin{tabular*}{.8\textwidth}{{@{\extracolsep\fill}cccccccccc}}
        \begin{tabularx}{\textwidth}{c c c *{8}{>{\centering\arraybackslash}X}}
		\toprule
		ACGAN                                          & Data     & \multicolumn{2}{c}{ResNet 50} & \multicolumn{2}{c}{ViT B 16} & \multicolumn{2}{c}{ViT L 16} & \multicolumn{2}{c}{LeViT Conv 384}                                                   \\
		Discriminator                                  & Type        & Acc.                           & F1                            & Acc.                          & F1                                 & Acc.  & F1    & Acc.           & F1             \\ \midrule
		- & only real & .7965       & .8800                          & .7952                         & .8802                         & .8005                              & .8833 & .7846 & .8713                           \\ \midrule
		ViT B 16, 4 clients                                  & only generated   & .3893                          & .4793                         & .7836                         & .8754                              & .5064 & .6263 & .6236          & .7514          \\
		                                             & generated + real & .7965                          & \textbf{.8810}                & .7734                         & .8644                              & .7737 & .8643 & .7750          & .8650          \\ %\cline{3-10}
		ViT B 16, 10 clients                                 & only generated   & .6291                          & .7464                         & \textbf{.8356}                & \textbf{.9100}                     & .7391 & .8398 & .7772          & .8672          \\
		                                               & generated + real & .7903                          & .8763                         & .7783                         & .8664                              & .7932 & .8790 & \textbf{.7968} & \textbf{.8809} \\
		\bottomrule
	\end{tabularx}
    % \end{adjustwidth}
\end{table*}

\begin{table}[tb]
    % \captionsetup{width=\textwidth}
    \caption{FID scores for data generated by GANs trained in the federated setting (lower is better). The GAN architecture used was ACGAN.}
	\label{tab:fid-fl}
	\centering
	\begin{tabular}{l c c}%*}{.5\textwidth}{@{\extracolsep\fill}l c c}
		\toprule
		Discriminator & No. of Clients & FID ($\downarrow$) \\
		\midrule
        ViT B 16 & 4  & 163.86 \\
                 & 10 & 181.50 \\
        SwinT    & 4  & 245.56 \\
                 & 10 & 167.22 \\
        \bottomrule
	\end{tabular}%*}
\end{table}

\subsubsection{Non-IID Datasets}

The fact that clients in the real world probably have non-IID data can be a significant problem, even affecting learning in some cases. To investigate how our model performs in this setting, we tried to simulate a non-IID data partitioning and compare it with the IID one.

To partition the training dataset, we split the real data, which is balanced in the original dataset, among six clients, ensuring that each client received a class distribution with a majority class amounting to $60$--$90\%$ of their data. Some clients had more MSIH samples, while others had more non-MSIH samples.

\begin{figure*}[tb]
	\centering
	\begin{subfigure}[b]{.32\textwidth}
		\includegraphics[width=\textwidth]{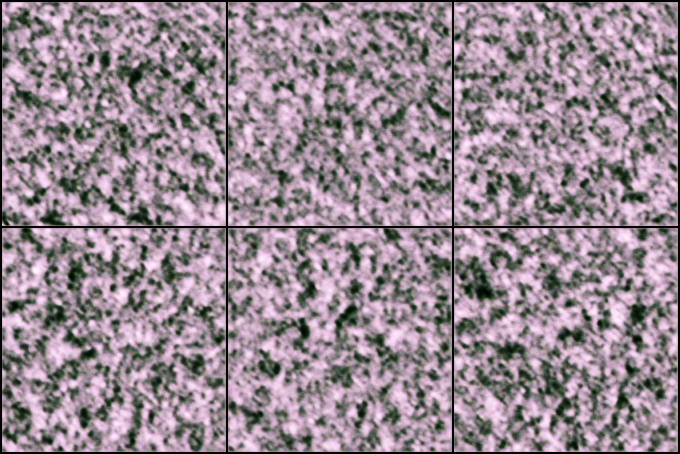}
		\caption{\centering After 1 epoch\vspace{8pt}}
	\end{subfigure}
	\begin{subfigure}[b]{.32\textwidth}
		\includegraphics[width=\textwidth]{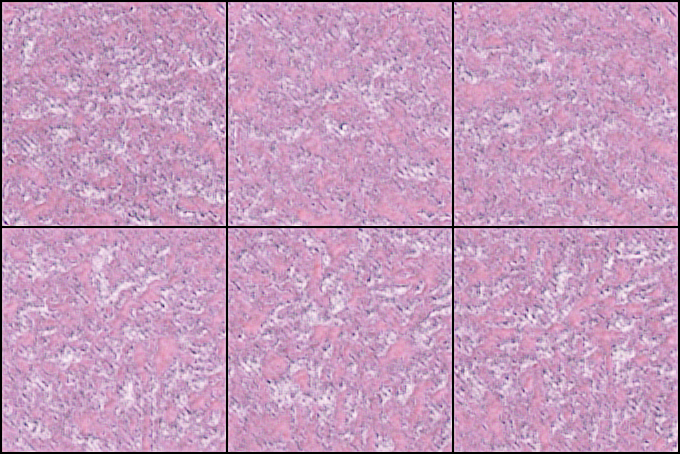}
		\caption{\centering After 2 epochs\vspace{8pt}}
	\end{subfigure}
	\begin{subfigure}[b]{.32\textwidth}
		\includegraphics[width=\textwidth]{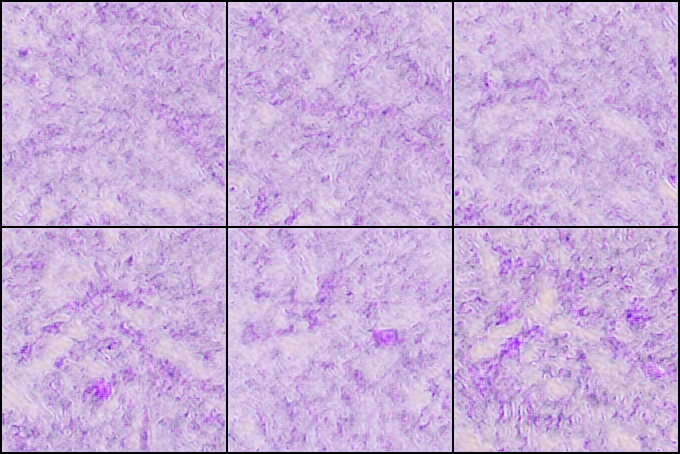}
		\caption{\centering After 4 epochs\vspace{8pt}}
	\end{subfigure}
	\begin{subfigure}[b]{.32\textwidth}
		\includegraphics[width=\textwidth]{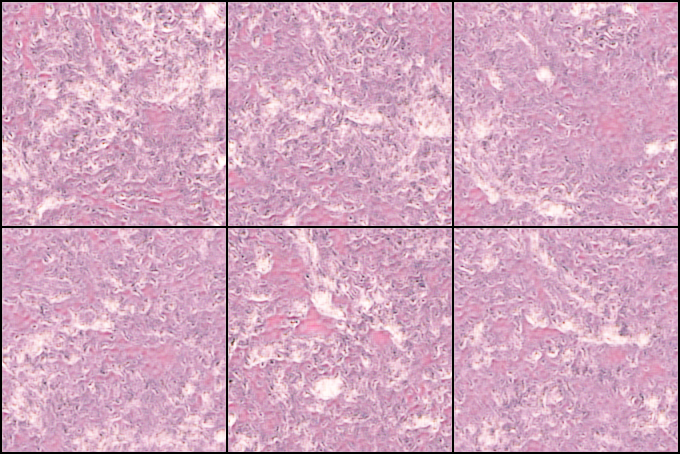}
		\caption{\centering After 6 epochs\vspace{8pt}}
	\end{subfigure}
	\begin{subfigure}[b]{.32\textwidth}
		\includegraphics[width=\textwidth]{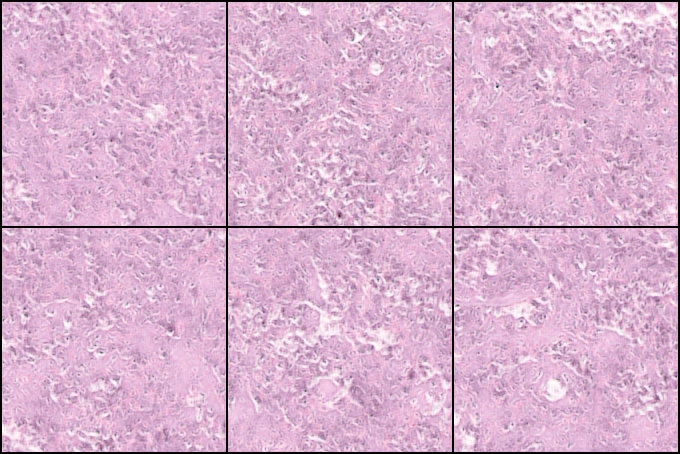}
		\caption{\centering After 8 epochs\vspace{8pt}}
	\end{subfigure}
	\begin{subfigure}[b]{.32\textwidth}
		\includegraphics[width=\textwidth]{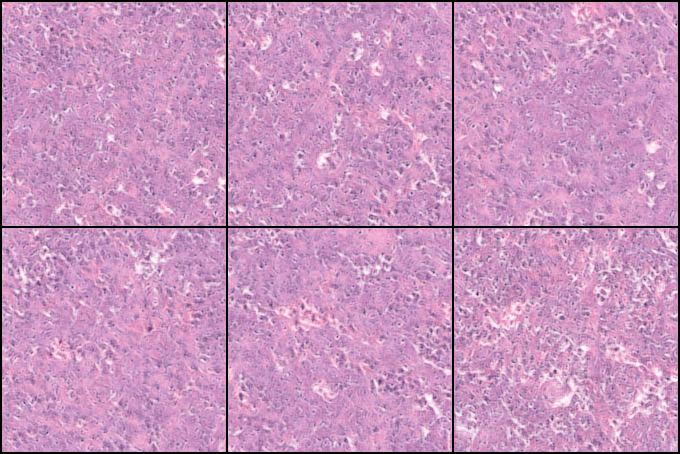}
		\caption{\centering After 12 epochs\vspace{8pt}}
	\end{subfigure}
	\caption{Examples of colonoscopy images when the non-IID ratio is $0.9$. Note that an epoch here refers to a round of training the server model. Each round iterates over the client's dataset multiple times.}
	\label{fig:non-iid}
\end{figure*}

On visual inspection, non-IID data did not seem to affect the quality of the final images for this task. However, it did affect the training time required to obtain images that looked reasonable. The more imbalanced configurations caused the training process to look more unstable in the beginning and required more epochs to start seeing good results. The evolution of training in one of these imbalanced settings can be seen in Figure \ref{fig:non-iid}.

The situation was similar for the downstream task. Images generated after training in the non-IID setting still improved the baseline quite often. We can see the corresponding scores in Table \ref{tab:augmented-classif-noniid}.

\begin{table*}[tb]
    % \begin{adjustwidth}{-\extralength}{0cm}
    \centering
	\caption{Classification scores when augmenting the training dataset with synthetic data from ACGAN + ViT, with different non-IID settings for FL clients. Improvements over the baseline (no synthetic data) are highlighted in bold.}
	\label{tab:augmented-classif-noniid}
% \begin{tabular*}{.8\textwidth}{{@{\extracolsep\fill}cccccccccc}}
\begin{tabularx}{\textwidth}{c c c *{8}{>{\centering\arraybackslash}X}}
\toprule
Non-IID               & Data                & \multicolumn{2}{c}{ResNet 50}  & \multicolumn{2}{c}{ViT B 16}   & \multicolumn{2}{c}{ViT L 16}   & \multicolumn{2}{c}{LeViT Conv 384} \\
Ratio (\%)            & Type                   & Acc.           & F1             & Acc.           & F1             & Acc.           & F1             & Acc.             & F1              \\
\midrule
- & only real & .7965          & .8800          & .7952          & .8802          & .8005          & .8833          & .7846            & .8713           \\
\midrule
60\%                    & only generated              & \textbf{.8215} & \textbf{.9017} & .7722          & .8697          & .7143          & .8288          & \textbf{.8180}   & .8996           \\
                      & generated + real            & .7787          & .8669          & .7889          & .8729          & .7914          & .8780          & .7755            & .8661           \\ %\cline{3-10} 
70\%                    & only generated              & \textbf{.8370} & \textbf{.9112} & \textbf{.8381} & \textbf{.9117} & \textbf{.8363} & \textbf{.9105} & \textbf{.8364}   & \textbf{.9108}  \\
                      & generated + real            & .7805          & .8699          & \textbf{.8023} & \textbf{.8850} & .7822          & .8691          & .7797            & .8686           \\ %\cline{3-10} 
80\%                    & only generated              & .2920          & .3255          & .3818          & .4605          & .2231          & .1503          & .1940            & .0856           \\
                      & generated + real            & \textbf{.8034} & \textbf{.8851} & .7941          & .8791          & .7652          & .8561          & .7351            & .8336           \\ %\cline{3-10} 
90\%                    & only generated              & \textbf{.9358} & .7450          & \textbf{.8019} & \textbf{.8896} & .7837          & .8753          & .3839            & .4596           \\
                      & generated + real            & .7895          & .8758          & .7931          & .8763          & .7917          & .8772          & \textbf{.7923}   & \textbf{.8774} \\
\bottomrule
\end{tabularx}
% \end{adjustwidth}
\end{table*}

\subsubsection{Production-grade Setup}

We implemented an experiment with a realistic setup to further test the feasibility of a federated learning system that spans multiple hospitals and allows them to keep their data private when training deep learning models. Using Kubernetes, we built a networked setup that used multiple actual nodes instead of simulating the network in software. Unlike an experimental setup, a more realistic one has to deal better with scalability, overhead issues, and security considerations.

\begin{figure*}[tb]
    \centering
    \includegraphics[width=\textwidth]{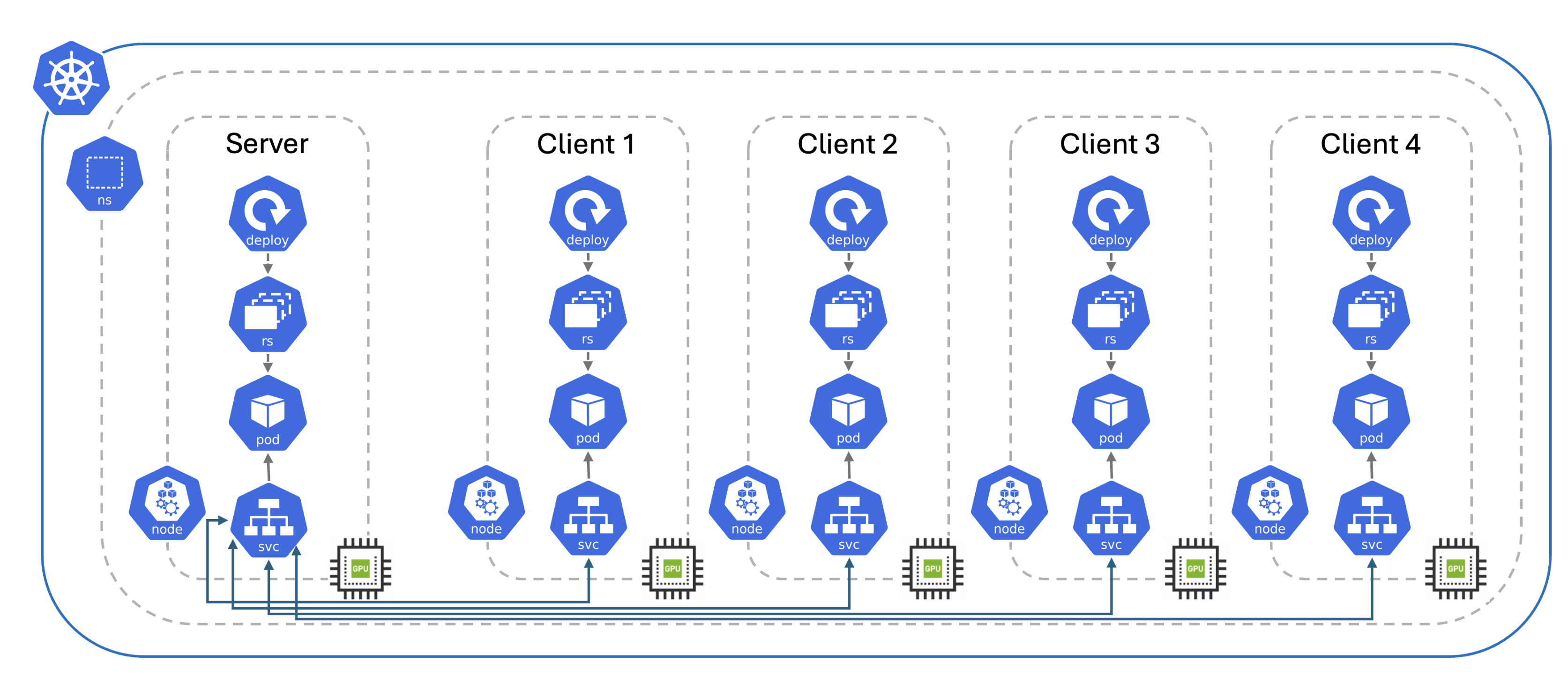}
    \caption{Kubernetes cluster architecture used for federated learning experiments.}
    \label{fig:k8s-arch}
\end{figure*}

Our implementation starts by creating a K8s cluster with a dedicated namespace that isolates our work from other tenants. We added GPU-enabled nodes to this cluster to support the high workload of running large machine learning models. For this experiment, we considered the case of four different hospitals (representing clients), and each of them was assigned to its node, which was used both to train the local version of the GAN and for communication with the server node. The aggregation server centralizes the model updates and coordinates training tasks throughout the network. A detailed diagram of this architecture can be seen in Figure \ref{fig:k8s-arch}.

The experiment was successful; we could use the entire cluster in parallel to train a version of the GAN with characteristics similar to those obtained in the federated setting described above.

%%%%%%%%%%%%%%%%%%%%%%%%%%%%%%%%%%%%%%%%%%
\section{Discussion}\label{sec:discussion}

\subsection{Did GANs Memorize the Training Data?}

\begin{figure*}[tb]
    \centering
    \includegraphics[width=0.5\textwidth]{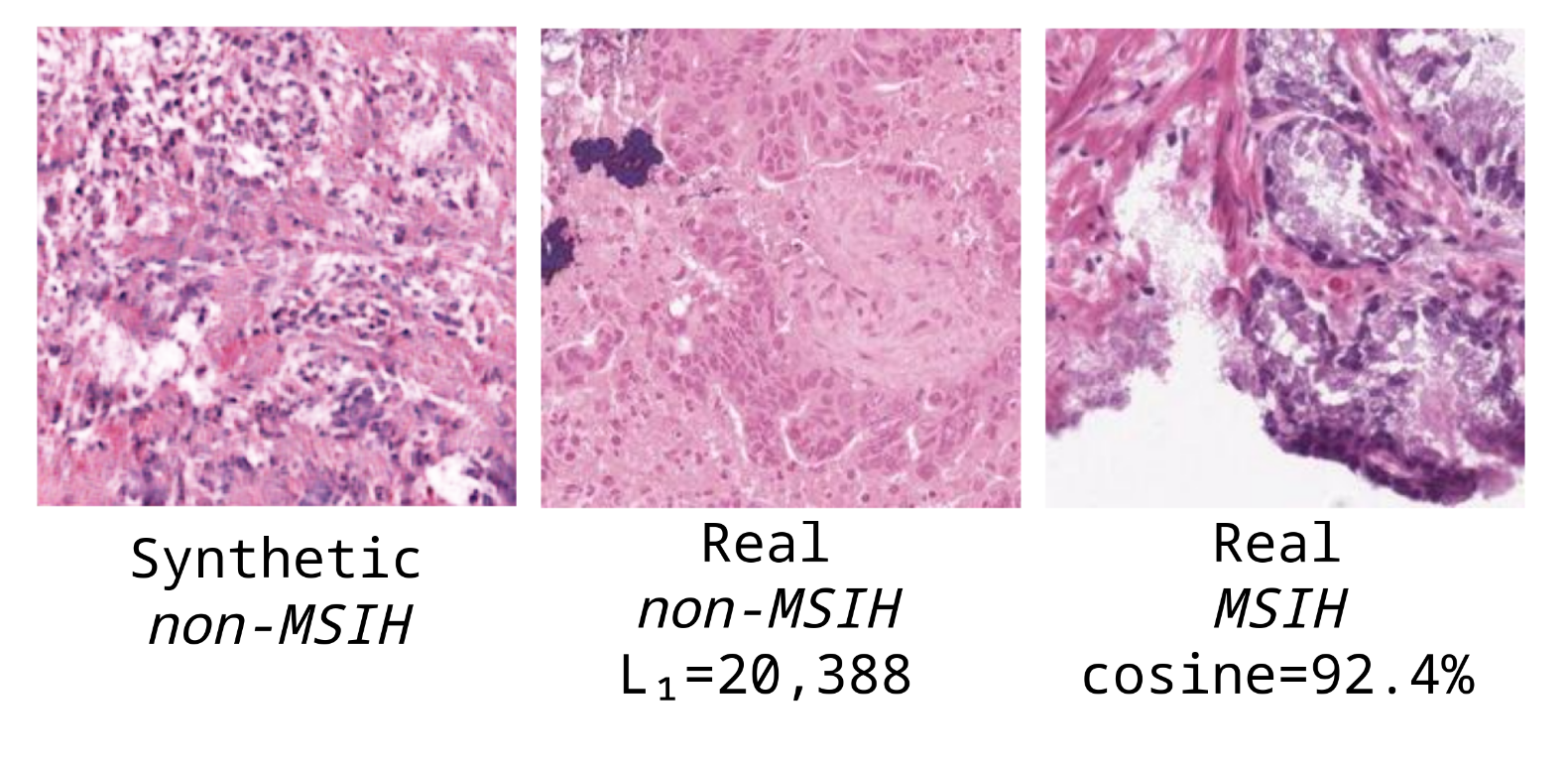}\\
    \includegraphics[width=0.5\textwidth]{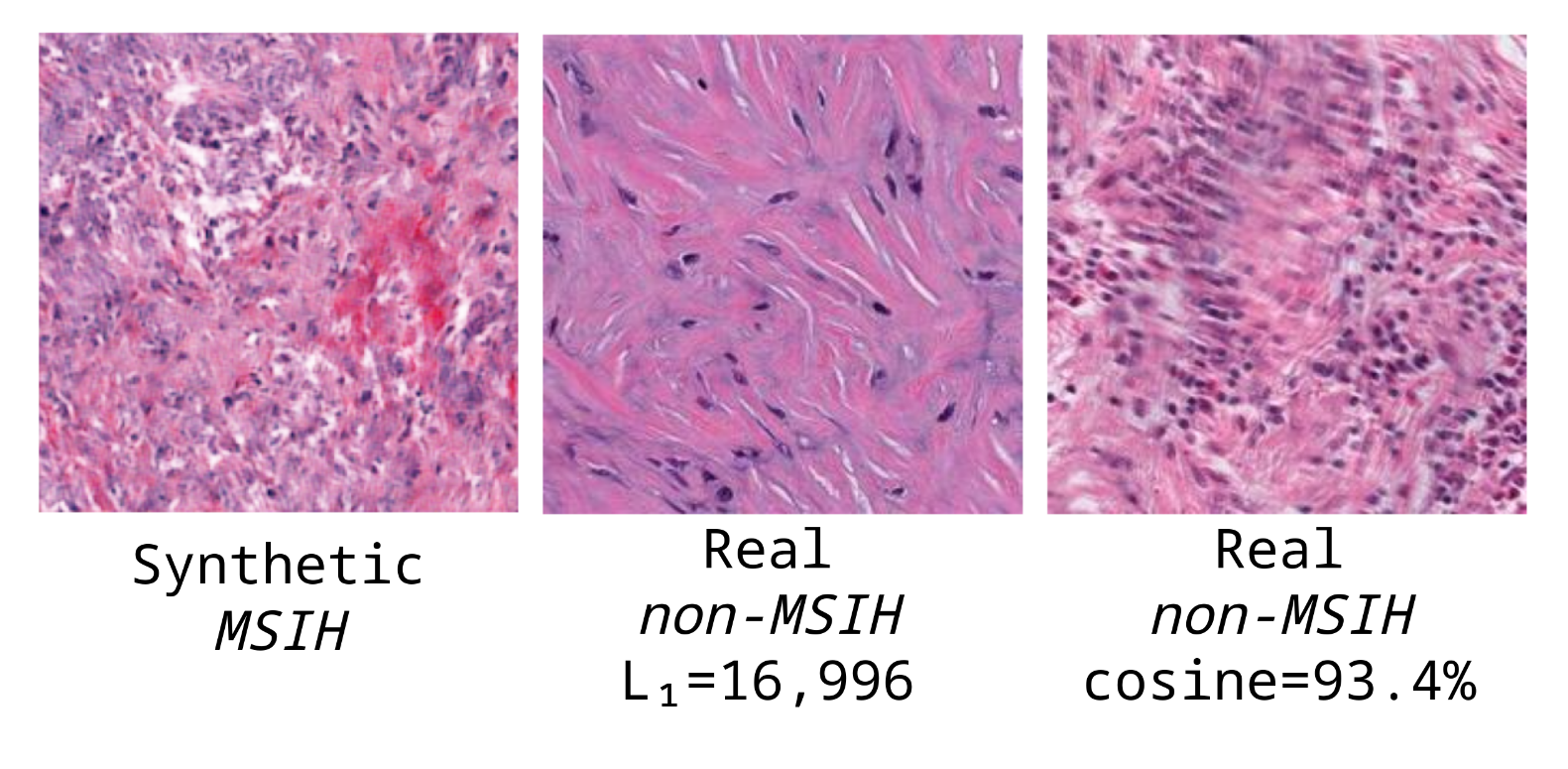}\\
    \includegraphics[width=0.5\textwidth]{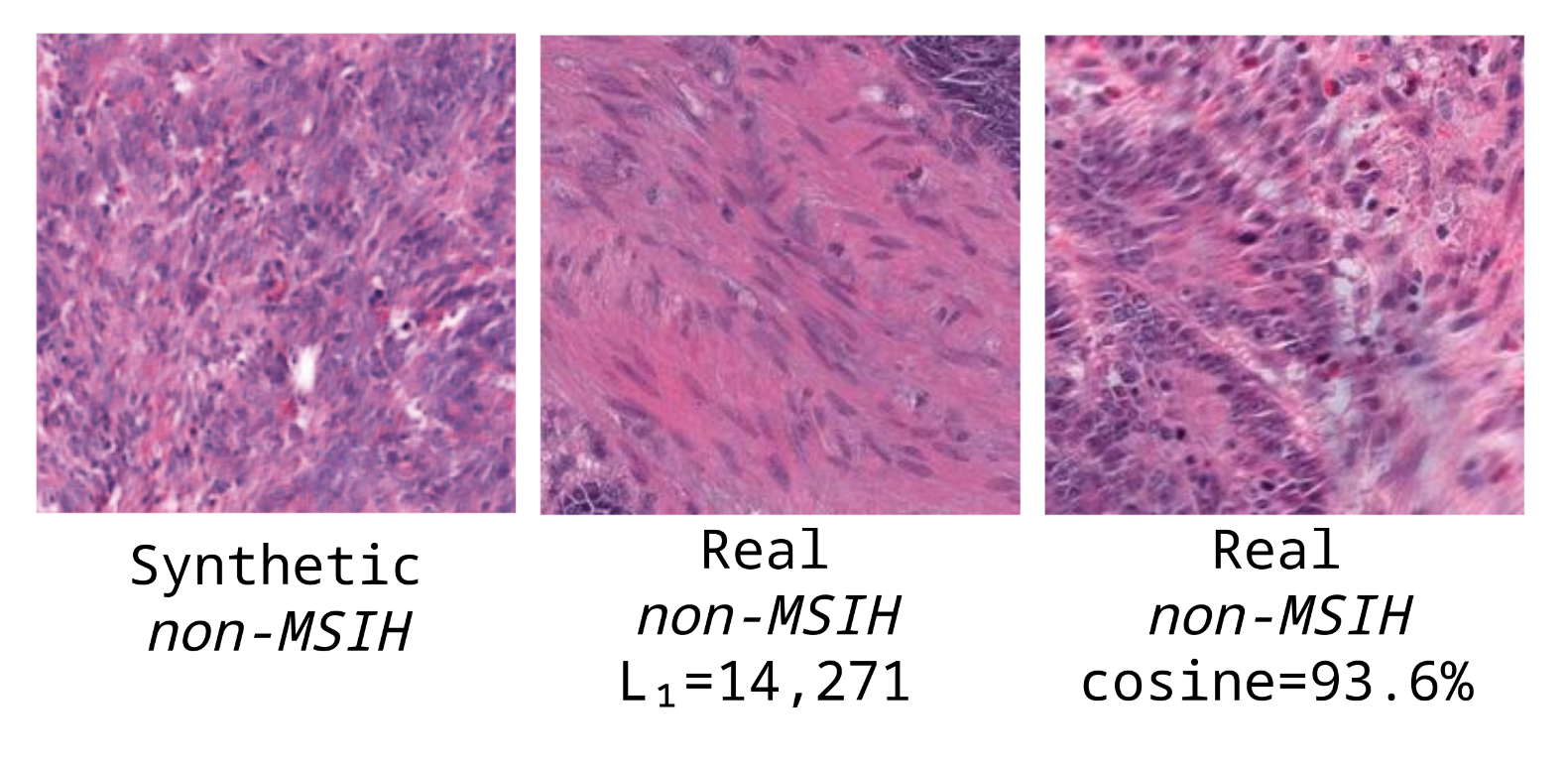}\\
    \includegraphics[width=0.5\textwidth]{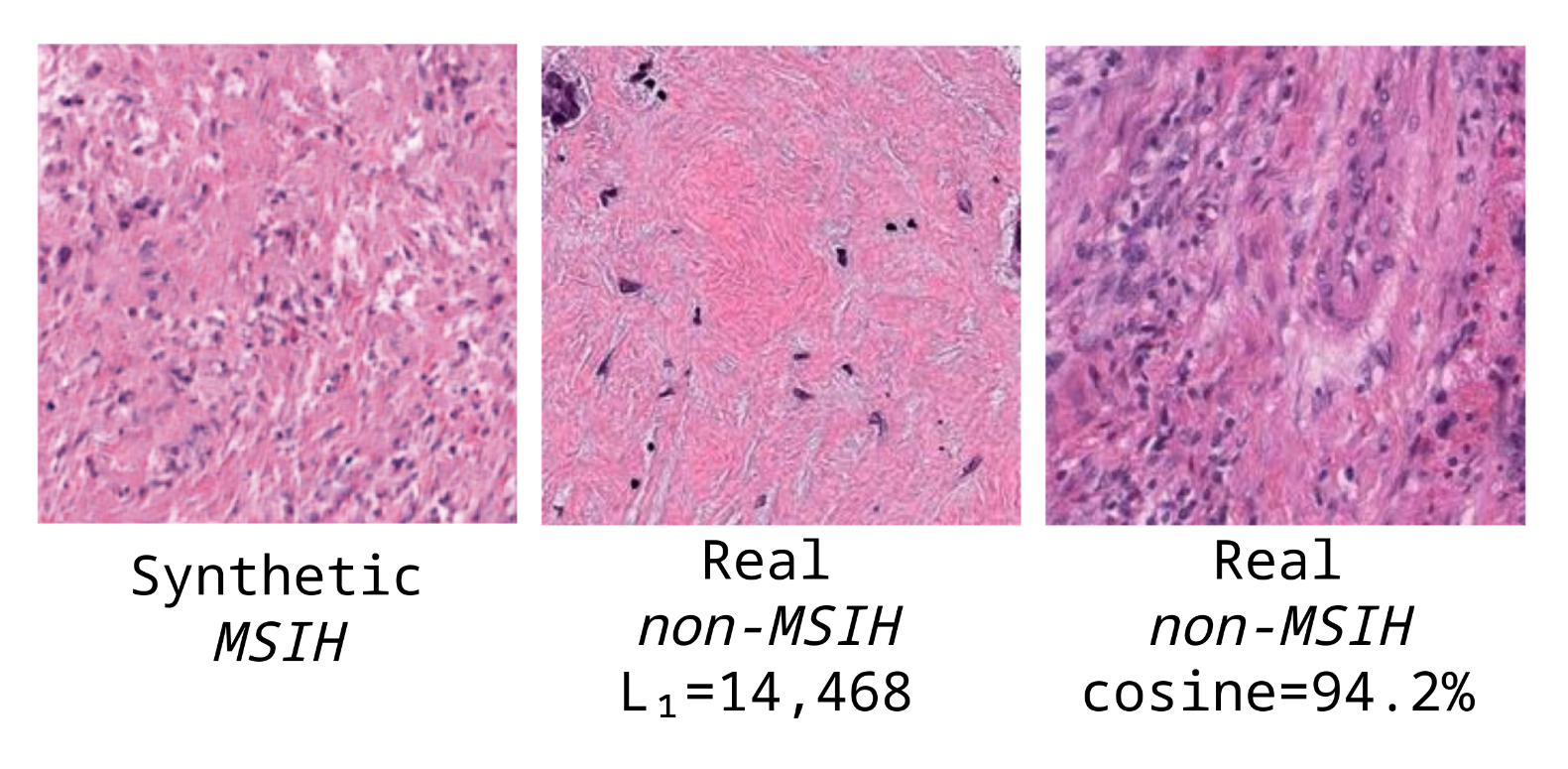}\\
    \caption{Comparison between four generated images and the training images closest to each. To measure similarity, we use the $L_1$ norm of the pixel-wise difference between the two images or the cosine similarity of the embeddings of the images. To compute embeddings, we use one of the classifiers trained on real images and treat one of its layers as a feature extractor.}
    \label{fig:fake-real}
\end{figure*}

The memorization of training data is an issue that GANs are in danger of experiencing. The generator could learn to fool the discriminator by producing replicas of the training images. This would obtain a small loss value, but would not create a useful model since it would be unable to generate novel samples. However, our experiments did not show signs of this problem. Not only does the loss value not indicate this, but Figure \ref{fig:fake-real} shows that the generated images are quite different from the ones in the training set.

\subsection{Did mode collapse occur in GANs?}

One common problem that GANs tend to experience is mode collapse, i.e., finding certain image types that manage to fool the discriminator and exploiting this by generating only images of those types. In our case, because the generator (i.e., CNN) and discriminator (i.e., ViT) fundamentally work differently, this helps to protect against mode collapse by introducing two different inductive biases. We can see that the model does not seem to experience mode collapse, generating many kinds for each image (for example, see Figures \ref{fig:centralized-vs-federated} and \ref{fig:fake-real}). However, we should note that the training set has much higher diversity, which is not reflected in the synthetic data.

\subsection{Class Similarity}

\begin{figure*}[tb]
	\centering
	\begin{subfigure}{.49\textwidth}
		\centering
		\includegraphics[width=\textwidth]{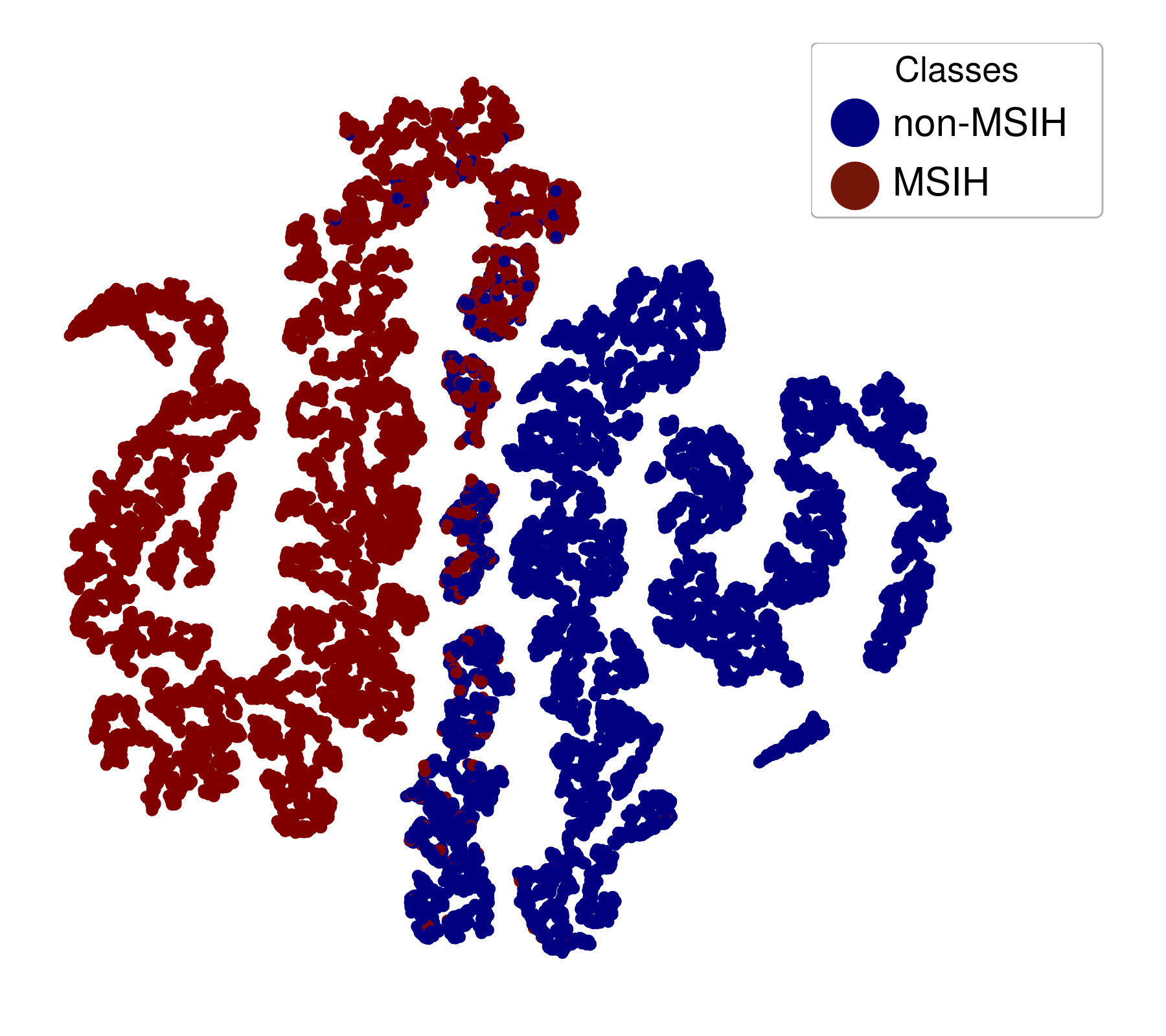}
		\caption{\centering Original Data}
	\end{subfigure}
	\hfill
	\begin{subfigure}{.49\textwidth}
		\centering
		\includegraphics[width=\textwidth]{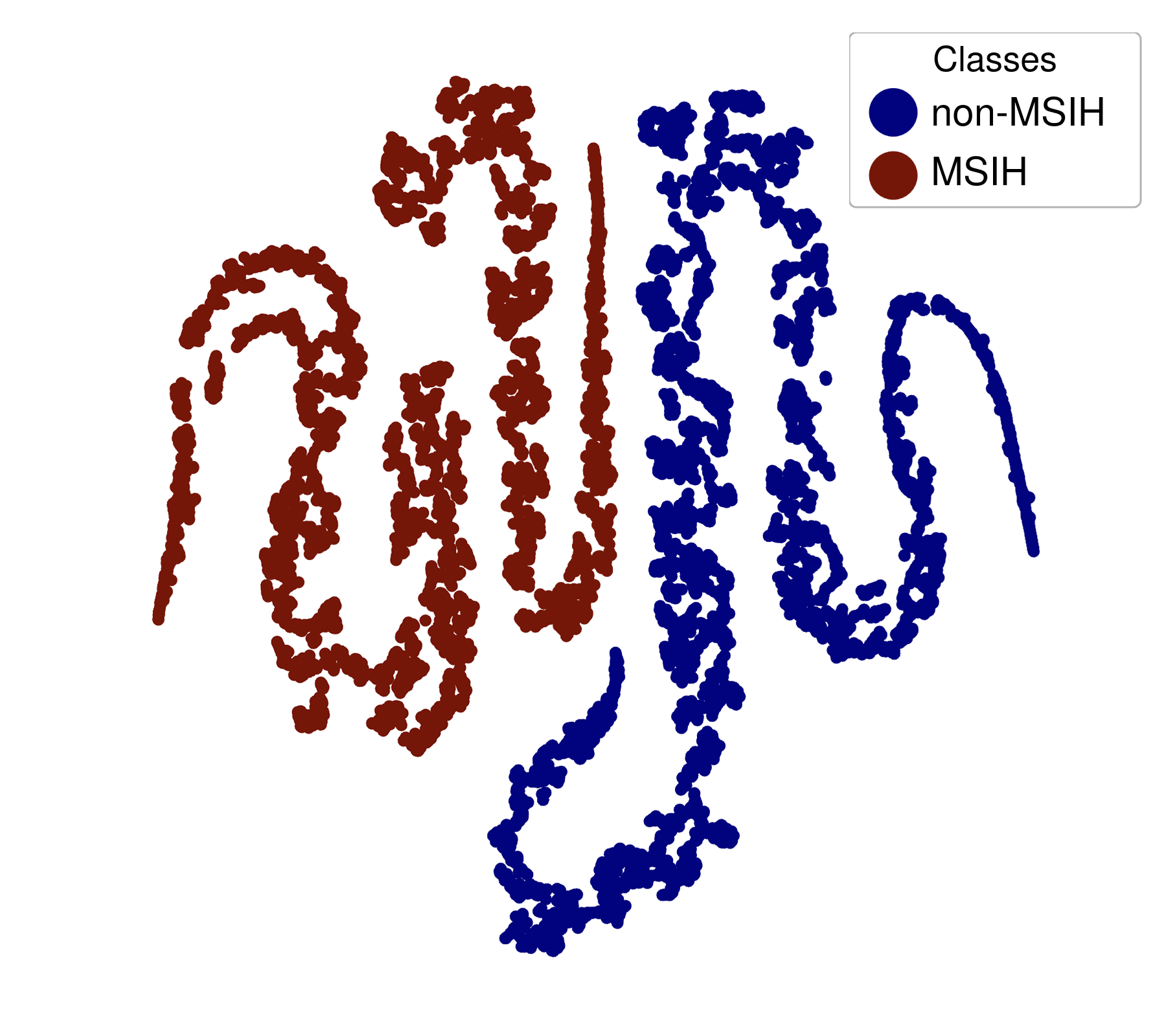}
		\caption{\centering Synthetic Data}
	\end{subfigure}
	\caption{Visualization of original and synthetic data using t‑SNE.}
	\label{fig:orig-vs-aug-tsne}
\end{figure*}

For our images to be useful for augmenting classification, we wish they encode some class-specific information. In addition to the classification scores presented before, we analyze them using data visualization. In Figure \ref{fig:orig-vs-aug-tsne}, we plot the embeddings produced by a ViT B 16 classifier for both real images and our synthetic images after reducing their dimension using the T-Distributed Stochastic Neighbor Embedding (t-SNE) technique~\citep{JMLR:v9:vandermaaten08a}. We can see that, while the two plots have similar shapes, our synthetic samples have much less overlap in the central area. This could signify a lack of images that are harder to classify and could teach a model something new.

% \begin{figure*}[tb]
%     \centering
%     \includegraphics[width=0.5\textwidth]{img/fake-vs-real1-newer.png}\\
%     \includegraphics[width=0.5\textwidth]{img/fake-vs-real3-newer.png}\\
%     \includegraphics[width=0.5\textwidth]{img/fake-vs-real2-newer.png}\\
%     \includegraphics[width=0.5\textwidth]{img/fake-vs-real4-newer.png}\\
%     \caption{Comparison between four generated images and the training images closest to each. To measure similarity, we use the $L_1$ norm of the pixel-wise difference between the two images or the cosine similarity of the embeddings of the images. To compute embeddings, we use one of the classifiers trained on real images and treat one of its layers as a feature extractor.}
%     \label{fig:fake-real}
% \end{figure*}

%%%%%%%%%%%%%%%%%%%%%%%%%%%%%%%%%%%%%%%%%%
\section{Conclusions}\label{sec:conclusion}

In this paper, we described the idea of augmenting classification datasets using synthetic data generated by GANs and experimented with it for a histological dataset. We compared multiple architectures and obtained promising results using ACGAN and ViT. We repeated experiments in the federated setting using Kubernetes as a production-ready framework. We concluded that this task and method are compatible with federated learning, which can have practical benefits in the medical context. Finally, we analyzed the images generated by ACGAN in more depth, searching for common problems that affect GANs, and observed that our method avoided some of these issues.

Regarding future research, it could be helpful to investigate a similar method for a multi-class classification task (as opposed to the binary classification tackled here). Similarly, we could improve the final performance of this dataset by using more complex networks as components in the GAN or by stabilizing the training process. Finally, diffusion models \citep{ho2020denoising} would constitute a different method of tackling the problem that has proven to be very effective and would merit comparison with the GAN approaches \cite{huang2023fastdiff}.

\section*{Limitations}

One limitation of our study is that, due to computational requirements, we did not perform extensive hyperparameter searches. The methods described here were susceptible to hyperparameters, so the results and conclusions could change somewhat after a more thorough search. Another limitation is that we restricted the experiments to a single binary classification dataset. The lack of more diverse datasets might limit the conclusions we can draw from our results.

\bibliographystyle{abbrvnat}  
\bibliography{references}  %%% Remove comment to use the external .bib file (using bibtex).

\end{document}